\pdfoutput=1
\documentclass[11pt]{article}

\usepackage[final]{acl}

\usepackage{times}
\usepackage{latexsym}

\usepackage[T1]{fontenc}
\usepackage[utf8]{inputenc}

\usepackage{microtype}

\usepackage{inconsolata}

\usepackage{graphicx}

\usepackage{multirow}

\title{Improving Causal Interventions in Amnesic Probing\\ with Mean Projection or LEACE}

\author{Alicja Dobrzeniecka\\
  NASK - \\ National Research Institute \\ Warsaw, Poland \\
  \texttt{alicja.dobrzeniecka@nask.pl} \\\And
  Antske Fokkens \and Pia Sommerauer \\
  Computational Linguistics and Text Mining Lab \\
  Dept. of Language, Literature and Communication\\
  Vrije Universiteit, Amsterdam, The Netherlands\\
  \texttt{antske.fokkens,pia.sommerauer@vu.nl} \\}

\begin{document}
\maketitle
\begin{abstract}
Amnesic probing is a technique used to examine the influence of specific linguistic information on the behaviour of a model. This involves identifying and removing the relevant information and then assessing whether the model’s performance on the main task changes. If the removed information is relevant, the model's performance should decline. The difficulty with this approach lies in removing \emph{only} the target information while leaving other information unchanged. It has been shown that Iterative Nullspace Projection (INLP), a widely used removal technique, introduces random modifications to representations when eliminating target information. We demonstrate that Mean Projection (MP) and LEACE, two proposed alternatives, remove information in a more targeted manner, thereby enhancing the potential for obtaining behavioural explanations through Amnesic Probing.
\end{abstract}

\section{Introduction}

A major paradigm for analyzing the linguistic information represented in language models is probing \cite{hupkes2018visualisation, belinkov2019analysis}. At its core, the probing paradigm assumes that the information captured by the model can be learnt from its intermediate representations. While this paradigm enables linguistically motivated analyses, it has several limitations, as summarised by \citet{Belinkov:2022}. 
A more recent framework of model analysis methods aims to establish \emph{causal} connections between specific linguistic properties and their impact on the model \cite{feder2022causal}. Amnesic Probing \cite{Elazar:etal:2021} tests whether a model has specific linguistic information by first \emph{identifying and removing} a property and then testing the impact of the removal on the behaviour of the model. \citet{Elazar:etal:2021} remove syntactic information and then test whether the model can still perform well on next-word prediction (a task that requires knowledge about syntax). If the syntactic properties were successfully identified and removed, the performance on the task should drop.  

A major factor in the Amnesic Probing framework is the \emph{removal} of the target property. \newcite{Elazar:etal:2021} use Iterative Nullspace Projection \cite[INLP]{ravfogel-etal-2020-null}, a method that has originally been proposed to debiase models. INLP removes information by iteratively applying projections to the embedding space. Each projection removes more of the target information.
A major risk of this method is removing information \emph{beyond} the target information and thus damaging other aspects of the vector space \cite{Haghighatkhah:etal:2022, zhang2022probing}.  
Multiple alternative methods that can remove information in a more targeted way have been proposed since \cite{ravfogel2022linear,ravfogel2022adversarial,Haghighatkhah:etal:2022,zhang2022probing,belrose2024leace}. \citet{belrose2024leace} prove the efficiency of these approaches. They also show that their LEACE approach indeed removes pos-information more effectively and distorts the overall space less than INLP hinting at LEACE being superior for amnesic probing. However, the impact of using alternative removal methods such as LEACE or Mean Projection (MP) has, to our knowledge, not been investigated in full Amnesic Probing experiments. Additionally, despite the limitations of INLP, it remains a popular approach.\footnote{We found that INLP received 125 citations and Elazar et al 59 citations of papers not mentioning LEACE or MP since 2024 on Google Scholar, including from papers using INLP. This shows that not all researchers using this method are aware of superior alternatives.} Therefore, it remains important to explain why alternative methods should be favoured.

%provide a first experiment on the impact of removing pos-information by their approach LEACE compared to INLP. They show that LEACE indeed removes pos-information more effectively and distorts the overall space less. %However, the impact alternative removal methods (such as LEACE or MP) on full Amnesic Probing experiments has not yet been investigated. 

%In this paper, we dive deeper into the impact of using a more targeted and efficient approach for removing target information in an amnesic probing setting. We provide a full and accessible outline of the amnesic probing experimental setup, adding additional checks and evaluations to the original experiments of \newcite{Elazar:etal:2021}.

This paper fills this gap and provides the following contributions. We provide an easy-to-follow step-by-step explanation of the Amnesic Probing framework, including additional checks and evaluations to the original experiments of \newcite{Elazar:etal:2021}. We systematically compare INLP against MP method proposed by \citet{Haghighatkhah:etal:2022} and \citet{zhang2022probing} and against LEACE on three control sets from \newcite{Elazar:etal:2021}. Our results demonstrate that, unlike INLP, both MP and LEACE improve the ability to uncover behavioural explanations via amnesic probing and successfully pass information control tests in two settings. Further analyses using our extended framework reveal that MP and LEACE more effectively remove the target property than INLP while causing less distortion to the remaining representation space. Although our original focus was on comparing INLP and MP — hence the more detailed treatment of MP — LEACE performs comparably to MP across all evaluations. We are also releasing a user-friendly framework.\footnote{\url{https://github.com/efemeryds/amnesic-probing-with-single-projection}}
% old link
%\url{https://anonymous.4open.science/r/amnesic-probing-with-single-projection-C57A}
%through INLP changes the space more while impacting the target behaviour less than MP.
    
%We expect to confirm prior results that showed that MP removes information in a more targeted manner and with less collateral damage compared to INLP \cite{Haghighatkhah:etal:2022, zhang2022probing}. We hypothesize that using MP improves the potential of obtaining behavioural explanations through Amnesic Probing. Our results confirm our hypothesis. MP outperforms INLP on the two control tests introduced by \citet{Elazar:etal:2021}. These results are in line with contemporary work by \citet{belrose2023leace} who show  there that linear concept erasure measure LEACE distorts the model less in amnesic probing settings compared to INLP. Our work differs from theirs in that we focus on the full procedure of applying amnesic probing, demonstrating the superiority of using MP over INLP in this line of work.

\section{Amnesic Probing}\label{sec.setup}

%Testing the impact of the removal on the embedding space is not trivial; we have to determine that (1) we removed the target property and (2) we did \emph{not} remove other information. The original framework proposed by \cite{Elazar:etal:2021} allows for such comparisons. In order to show the impact of the removal methods on the vector space and the conclusions drawn in probing experiments, we closely follow the original amnesic probing framework (ref). 

%\pia{Using Antske's summary and syntax example to convey the idea here:} 

The Amnesic Probing method, as proposed by \citet{Elazar:etal:2021}, is based on the following idea. If we want to know, for example, whether a specific model uses syntactic information to predict the next word, we test the following:
(1) whether the model contains syntactic information;
(2) whether it performs worse at predicting the next word after the syntactic information is removed;
(3) whether this drop in performance is greater than that observed in models with the same amount of random information removed;
(4) whether explicitly adding syntactic information brings performance close to that of the original model. First, we provide an overview of the experimental setup and then describe the intuition behind the removal procedures used in this study.

\subsection{General Experimental Set-Up}

%We follow the procedure of \citet{Elazar:etal:2021} closely. 
We motivate the general idea here, closely following \citet{Elazar:etal:2021}, and provide a more detailed description of our experimental setup and results in Section~\ref{sec:experimentresults}. We first report reproduced INLP results that are not identical, but tell the same overall story as in Elazar et al. We hypothesize that Amnesic Probing will yield clearer results when MP or LEACE is used instead of INLP since it will remove target information more precisely. In particular, we expect that (1) MP and LEACE will lead to a larger performance drop than models modified by the same number of random projections (information control). (2) We expect that the original performance on the task can be well restored by adding explicit gold labels to the modified vector space (selectivity control). 

%We briefly describe the target information and main task and explain the two removal methods we compare. A detailed description of our experimental setup and results are provided in Section~\ref{sec:experimentresults}.
%We report INLP results obtained by replicating Elazar et al.'s original work. Our INLP results are not identical but they do tell the same overall story.

\citet{Elazar:etal:2021} study the impact of 6 linguistic properties: \textit{syntactic dependency (dep)}, \textit{f-pos}, \textit{c-pos}, \textit{named-entity (ner)}, \textit{phrase start} and \textit{phrase end}. Our experiments cover the properties \textit{dep}, \textit{f-pos} and \textit{c-pos} from the Universal Dependencies (UD) English dataset \citep{mcdonald2013universal}. 
Removing these properties had a clear impact on performance in the main task in \citet{Elazar:etal:2021} and the original data splits were available for reproduction. Data was not available for the other three tasks and INLP did not yield clear results for them. Including these additional experiments would likely confirm that neither INLP, MP, nor LEACE is effective in this context. Due to the lack of data and the limited expected insights, we chose not to include these experiments.

%whereas the other three did not. Since we do not expect more efficient removal to have more absolute impact on the main task compared to removal with more collateral damage, adding these experiments is unlikely to lead to more insights. Moreover, reproducing the original results was not possible, because the data was not available. %We left thus them out.

%whereas the others did not. We do not expect a more efficient removal method to have a larger absolute impact on the main task, adding these experiments is unlikely to lead to new insights. 

%Moreover, we could obtain the linguistic properties from the Universal Dependencies (UD) English dataset \citep{mcdonald2013universal}, whereas the original data for the ner experiments were not available.

\citet{Elazar:etal:2021} use two metrics to test the impact of the information they removed. First, they evaluate performance on language modeling as a main task. Here they measure next-word prediction accuracy. Second, they calculate the Kullback-Leibler Divergence ($D_{KL}$), a useful measure to investigate the difference between two probability distributions. 

% In this case, we present $D_{KL}$ between the erased models and their vanilla baselines for language modeling with non-masked BERT.

%The $D_{KL}$ of the distribution \textit{q(x)} from \textit{p(x)} checks how much information is lost when the \textit{p(x)} distribution is approximated by \textit{q(x)}. It can be used to measure the impact on the overall distribution after transformation.

\subsection{INLP, MP and LEACE}\label{sec:projectionmethods}

The effectiveness of Amnesic Probing depends entirely on the quality of the method used to remove information. If the method is not precise enough, it becomes impossible to determine whether the drop is due to the removal of target information or loss of other information.

%removed: (initially proposed by ravf (maybe put back with more space
\paragraph{INLP}
\citet{Elazar:etal:2021} used INLP \citep{ravfogel-etal-2020-null} in their amnesic probing setup. INLP removes information as follows: a linear SVM model is trained to distinguish the classes of the target information. An orthogonal projection is applied along the decision boundary to \emph{remove} the target information for each class. This process is repeated $n$ times or until the classification performance remains below a threshold. This approach requires a relatively high number of projections to remove the target information: in \citet{Elazar:etal:2021} $n$ is set to 20, meaning that for target information with 41 classes, we can end up with a maximum of 820 projections. This relatively high number of projections is seen as the main cause that INLP removes other information beyond the target information \cite{Haghighatkhah:etal:2022, zhang2022probing}. 

\paragraph{MP} \citet{Haghighatkhah:etal:2022,zhang2022probing} propose an alternative method that does not require multiple iterative projections. Instead of training a model on the target property, this method simply takes the mean of the data points in each class to find the directions for their projections. This is a more efficient way to find an optimal projection compared to learning-based approaches, particularly when class distributions are unbalanced \citep{Haghighatkhah:etal:2022}. Unlike INLP, MP requires only one projection per class and the number of directions to be removed is stable and equal to the number of target classes. This can significantly reduce its detrimental effect on the rest of the representation. \citet{zhang2022probing} confirm that MP removes syntactic information in a more targeted way than INLP. We create the set of directions in the same way as \citet{zhang2022probing}: \textit{$w_{i}$} = \textit{$u_{i}$} - \textit{$u_r$}, where \textit{$u_{i}$} is for the mean of a current class and \textit{$u_r$} stands for the total mean of all the remaining classes excluding \textit{$u_{i}$}.

\paragraph{LEACE}
The LEACE method, proposed in \citet{belrose2024leace}, removes target information in a more precise manner, similarly to MP. Both LEACE and MP perform linear projection-based concept erasure using label-driven subspaces. However, LEACE is more theoretically grounded and proves the removal of all linear information about the concept. Both methods have their pros and cons: LEACE provides more theoretical guarantees, while MP is heuristic. At the same time, LEACE is more computationally intensive, whereas MP is simple and straightforward to apply. %This paper presents the results of both methods, but focuses primarily on MP.

%We follow the procedure of \citet{zhang2022probing} and create the set of directions as follows:

% \citet{zhang2022probing} confirm that MP removes syntactic information in a more targeted way than INLP. 

\section{Experiments and Results}\label{sec:experimentresults}

We walk through the amnesic probing procedure step-by-step and present the results of our comparison of INLP, MP and LEACE. \citet{Elazar:etal:2021} conducted experiments on two variants of the BERT model (masked and unmasked) \cite{devlin2019bert}. Both led to the same insights. We present the results for unmasked BERT here, and for masked BERT in Appendix~\ref{appendix:masked_setup}. 

%In this section, we walk through the amnesic probing procedure step-by-step and present the results of comparing INLP and MP. \citet{Elazar:etal:2021} present experiments on two model-variants (masked and unmasked) of BERT \cite{devlin2019bert}. Both led to the same insights, we show results for unmasked BERT here, and for masked BERT (including results for LEACE) in Appendix~\ref{appendix:masked_setup}. LEACE results on unmasked BERT are provided in Appendix~\ref{app:results_leace}.

\paragraph{Step 1: Can we identify and remove target information?}

\begin{table}[hb]
\centering
\setlength{\tabcolsep}{6pt} 
\renewcommand{\arraystretch}{1.3}
\begin{tabular}{ c c c c c }
\hline
&  & \textit{dep} & \textit{f-pos} & \textit{c-pos} \\
\hline
& N. classes & 41 & 45 & 12  \\ 
& Majority class \% & 11.44 & 13.22 & 31.76 \\
\hline
& Vanilla & 76.77 & 90.04 & 92.44 \\ 
& Vanilla (LEACE) & 74.65 & 89.03 & 92.01 \\ 
& Amnesic INLP & \textbf{9.73} & \textbf{10.79} & 33.91  \\
& Amnesic MP & 14.33 & 16.66 & \textbf{24.06} \\
& Amnesic LEACE & 14.33 & 16.28 & 25.87 \\ 
\hline
\end{tabular}
\caption{Results for probing for non-masked BERT encoding. Values in bold indicate the largest change among the same properties. Probing on Vanilla was repeated when adding results for LEACE.}
\label{table:probing_non_masked}
\end{table}

%The first step in the process entails \textbf{successfully removing the target information}. 
Following \citet{Elazar:etal:2021}, we first verify whether the target information is present in the original model through a regular probing procedure. We test whether a linear classifier can learn the target information from the vanilla model (i.e.\ in its original state). We then apply INLP as well as MP and LEACE to create alternative models where target information is removed. We add an additional test compared to \newcite{Elazar:etal:2021} and check whether probing indeed fails on the target property after applying INLP and MP. The results are presented in Table~\ref{table:probing_non_masked}. INLP, MP and LEACE all lead to a large drop in accuracy for probing the target information. For \textit{dep} and \textit{f-pos} the model performs below the majority class after applying INLP and slightly above it when MP or LEACE is applied. For \textit{c-pos}, MP and LEACE bring the model down to performance below majority class and INLP remains slightly above. MP and LEACE yield highly comparable performance: MP does slightly better for \textit{f-pos}, LEACE for \textit{c-pos}.

We add two comparisons to investigate the impact of each method on the overall model: (1) the change in matrix rank \cite{Strang2015IntroLinear} and (2) cosine similarity between layers before and after modification. Both tests show that INLP has a much larger impact on the model than MP and LEACE. A detailed description of these metrics and the results can be found in Appendix~\ref{app:modelrankcosine}.

%larger impact on the model as a whole compared to MP

% shorter alternative
% Probing the target property is highly impacted by both INLP and MP, though INLP causes a larger drop in result for \textit{dep} and \textit{f-pos} and MP for \textit{c-pos}.

\paragraph{Step 2: What is the impact on model behaviour?}
 
 We verify whether removing the target information has an impact on the main task. We first compare the vanilla model's performance with the modified model's performance to see if removing the information leads to a decrease. If this is not the case, the information we removed did not provide support to carry out the task in the original model.

 %\textbf{779} & \textbf{765} & \textbf{240}
  %for reasons of space

%The number of random directions is equal to the number of directions removed by the method we are comparing to. For INLP we apply 779 projections for \textit{dep}, 765 for \textit{f-pos} and 240 for \textit{c-pos}. For MP, this amounts to 41, 45 and 12 projections.\footnote{We also apply a Dropout baseline, where we randomly select columns in the embedding space and replace them with $0s$. The number of columns replaced is equal to the number of directions removed by the method. Dropout yields the same results as the random baseline, so we omitted it for reasons of space).} 

%he results for LEACE are highly comparable to MP, with the results for probing dropping less than those obtained when running INLP for \textit{dep} and \textit{f-pos}, but more for \textit{c-pos}. The end results for \textit{f-pos} are slightly worse when applying LEACE and for \textit{c-pos} slightly better compared to MP. 
%Models where MP was used for removing information, on the other hand, consistently lead to a larger drop compared to cases where the same number of random projections were used.

%Appendix~\ref{appendix:category} provides the results of an analysis testing the impact of distortions for specific cpos categories. Like \citet{Elazar:etal:2021} we find that distortions mainly affect function words.

 However, a drop in performance does not automatically mean that the target information helped. The drop can also be caused by other aspects of the modification. We compare the effect of removing target information with the impact of removing a comparable amount of random information (information control). If the target information is relevant, the removal of the target information should have a \emph{higher negative impact} than the removal of random information. 
 %removing this information should have more negative impact on the task compared to removing random information. 
The Amnesic Probing framework contains two baselines that apply random modifications to the embedding space: For a Random baseline, we apply a projection with randomly chosen directions to the embedding space. The number of random directions is equal to the number of directions removed by the method we are comparing to. For INLP we apply 779 projections for \textit{dep}, 765 for \textit{f-pos} and 240 for \textit{c-pos}. For MP, this amounts to 41, 45 and 12 projections and for LEACE to 41, 45 and, 12 respectively.\footnote{We also apply a Dropout baseline, where we randomly select columns in the embedding space and replace them with $0s$. The number of columns replaced is equal to the number of directions removed by the method. Dropout yields the same results as the random baseline, so we omitted it to save space.}

\begin{table}[h!]
\centering
\scalebox{0.9}{
\setlength{\tabcolsep}{6pt} 
\renewcommand{\arraystretch}{1.3}
\begin{tabular}{ l l c c c}
\hline
& & \textit{dep} & \textit{f-pos} & \textit{c-pos} \\
\hline
\multirow{9}{*}{Acc} & Vanilla & 94.12 & 94.12 & 94.12 \\
& Amn. INLP (repr) & 8.74 & 14.6 & \textbf{72.95}  \\ 
& Rand. INLP (repr) & 4.67 & 4.86 & 90.21  \\ 
& Amn. MP & \textbf{79.49} & \textbf{65.39} & \textbf{86.87} \\ 
& Rand. MP & 93.98 & 93.85 & 94.05 \\
& Amn. LEACE & \textbf{86.82} & \textbf{70.96} & \textbf{90.18} \\
& Rand. LEACE & 93.71 & 94.01 & 93.95 \\
%\hline
%\multicolumn{2}{c}{Amn. INLP (Elazar)} & \textbf{7.05} & \textbf{12.31} & \textbf{61.92} \\
& Amn. INLP (Elazar) & \textbf{7.05} & \textbf{12.31} & \textbf{61.92} \\
& Rand. INLP (Elazar) & 12.31 & 56.47 & 89.65  \\ 
\hline

\multirow{8}{*}{$D_{kl}$} & Rand. INLP (repr) & 8.98 & 8.9 & 0.47 \\ 
& Rand. MP  & 0.28 & 0.02 & 0.0 \\
& Rand. LEACE & 0.02 & 0.02 & 0.00 \\
& Amn. INLP (repr) & 7.97 & 6.08 & \textbf{1.99}  \\ 
& Amn. MP & \textbf{0.89} & \textbf{1.63} & \textbf{0.49} \\
& Amn. LEACE & \textbf{0.34} & \textbf{1.06} & \textbf{0.19} \\
%\hline
 & Rand. INLP (Elazar) & 8.11  &	4.61  &	0.36 \\
& Amn. INLP (Elazar)  & 	\textbf{8.53} &	\textbf{7.63} &	\textbf{3.21}
\end{tabular}}
\caption{LM task for non-masked BERT encoding comparing results of Amnesic (Amn) probing methods and the impact of the same number of random projection (Rand). Bold values reflect scenarios where the amnesic value had a bigger impact on accuracy (Acc) or led to a lower change in general distribution of tokens ($D_{kl}$) compared to random projections. We include the results reported by \newcite{Elazar:etal:2021} (Elazar) for comparison to our reproduced results (repr).}
\label{table:lm_non_masked}
\end{table}

Table~\ref{table:lm_non_masked} provides the results on language model accuracy and calculating $D_{KL}$ for all models and their baselines. Applying INLP leads to a much larger drop in accuracy for the main task and a much higher $D_{KL}$ compared to MP and LEACE. When INLP is used to remove dependencies or f-pos, more damage is done by applying the same amount of random projections in our experiments. 
%This shows that Amnesic Probing fails to prove that this information indeed impacted the main task when using INLP. 
This check shows that we cannot establish a causal link between the intervention of removing information with INLP and model behaviour on the main task. When MP or LEACE is used, we can establish this link; here, we can consistently observe a higher impact of the target information removal by MP compared to random removal.

Table~\ref{table:lm_non_masked} also includes the original results reported by \newcite{Elazar:etal:2021}. Results from INLP are influenced by random factors and in our reproduced study more projections were needed (see Limitations). \newcite{Elazar:etal:2021} obtained meaningful for \textit{f-pos}, next to \textit{c-pos}. However, the results for MP and LEACE still appear to be much more precise than those reported by \newcite{Elazar:etal:2021}. Although the same number of random projections as in the original INLP study led to a significant drop in results compared to the vanilla model, the corresponding amount of random projections applied by MP or LEACE hardly affects the results. 

%
%Table~\ref{table:lm_non_masked_with_leace} also provides the original results as reported by \newcite{Elazar:etal:2021}. Their results are also meaningful for \textit{f-pos}. However, here as well, the results for MP and LEACE appear to be much more precise. The same number of random projections barely impacts the model's performance, which clearly demonstrates that the impact on performance is related to the information provided by the linguistic property. Because the same number of random projections already has a significant impact in the case of INLP's original results, it remains unclear what part of the drop in performance is purely due to contributions by \textit{f-pos} and what is the result of loosing \textit{f-pos} information in a model that is generally damaged by the projections applied to it.

%\section{Category Analysis Results}\label{appendix:category}

\begin{table}[htbp]
\scalebox{0.90}{
\setlength{\tabcolsep}{6pt} 
\renewcommand{\arraystretch}{1.6}
\begin{tabular}{ c c c c c}
\hline 
Cat&Vanilla&$\delta$ INLP&$\delta$ MP &$\delta$ LEACE\\
\hline
ADJ&16.67&0.00&-4.17 & 0.00\\ 
NOUN&53.85&16.24&5.13 & -0.85\\
ADP&56.52&21.74&0.00 & -8.70\\
DET&88.24&79.41&67.65 & 67.65\\
CONJ&80.00&80.00&80.00 & 60.00\\
VERB&42.31&9.62&0.00 & 0.00\\
ADV&54.55&18.18&0.00 & 18.18\\
.&74.47&31.91&2.13 & 4.26\\
PRT&100.00&44.44&0.00 & 0.00\\
-&100.00&0.00&0.00 & 0.00\\
PRON&100.00&50.00&50.00 & 50.00\\
NUM&22.22&-11.11&0.00 & -11.11\\
\hline
\end{tabular}
}
\caption{Analysis of the impact of the distortion per c-pos category. $\delta$ INLP/MP/LEACE show the drop in results after projections. Negative values represent increased results.}
\label{table:cpos_cat}
\end{table}

Table~\ref{table:cpos_cat} provides the results of an analysis testing the impact of distortions for specific cpos categories, in line with the analysis carried out by \newcite{Elazar:etal:2021}. It presents the accuracy of the original model, and the drop in accuracy after applying INLP, MP or LEACE for each pos-category. We observe that all projection methods mainly impact function words (conjunction, determiners and pronouns). This difference is even more pronounced when using MP compared to INLP. MP and LEACE have less impact on nouns than INLP and no impact (MP) or an improvement (LEACE) on accuracy of predicting verbs and adpositions, whereas INLP does impact these negatively. MP and LEACE do not appear to have an impact on predicting particles (which are also function words), INLP does impact their prediction.

% \paragraph{Step 3: Is the performance drop caused by the removal of the target information or by damaging the vector space?}
 %In order to establish whether it is indeed the target information that helped the model two verification steps are carried out. 
 
% \textbf{We compare the effect of removing target information to the impact of removing a comparable amount of random information}. If the target information is relevant, the removal of the target information should have a \emph{higher negative impact} than the removal of random information. 
 %removing this information should have more negative impact on the task compared to removing random information. 
%The amnesic probing framework contains two baselines that apply random modifications to the space: \pia{From Alicja:} For a Random baseline, we apply a projection with randomly chosen directions to the embedding space. The number of random directions is equal to the number of directions removed by the method. \pia{From Alicja, but modified} The same reasoning applies to a Dropout baseline. Here we randomly select columns in the embedding space and replace them with $0s$. The number of columns replaced is equal to the number of directions removed by the method.

 %\pia{Results in Table~\ref{table:lm_non_masked}}

 %\vspace{-0.3cm}
 
\paragraph{Step 3: Can the performance be restored by putting the target information back into the vector space?}
 We carry out the selectivity control test and see whether the performance on the main task can be restored by adding explicit gold labels of the target information to the modified space. If the target information contributed to the original result and the loss in performance was mostly due to this information now being absent, performance should be largely restored. In contrast, if the drop in performance was caused by other information that was removed, these gold labels will not be able to recover for the effect. 
 We present the results in Table~\ref{table:selectivity}.\footnote{Note that the modification needed to add gold syntactic information requires us to remove the sequential container. This explains the, sometimes substantial, drop in results of the amnesic models in Table~\ref{table:selectivity} compared to Table~\ref{table:lm_non_masked}.} When adding gold labels, results for INLP, MP and LEACE all improve showing that the target information was responsible for the drop in performance at least to some extent. As expected, the results approximate the original result much better when MP or LEACE is used, coming close to the original result again. In INLP, we see that loss of other information also played a role. This confirms that MP and LEACE removed the target information more precisely and the results obtained while using MP provide a more reliable insight into the extent to which the target information supported the task.

%
%Table~\ref{table:selectivity_with_leace} provides the results for the selectivity experiments including LEACE. The results for LEACE are in line with those of MP, with slightly higher differences for LEACE. 
 
 % Second, we carry out what \citet{Elazar:etal:2021} call a  selectivity control experiment and test to whether \textbf{adding gold target labels largely restores} original performance of the model. 
 %what happens when we provide the target information directly to the modified model through gold labels. If the target information supports the task, performance on the main task should be largely restored to the result on the original model.

 \begin{table}[htbp]
\centering
% \scalebox{1.3}{
\setlength{\tabcolsep}{6pt} 
\renewcommand{\arraystretch}{1.3}
\begin{tabular}{ c c c c c}
\hline
& \textit{dep} & \textit{f-pos} & \textit{c-pos} \\
\hline
Amnesic INLP & 5.92 & 4.95 & 0.49 \\ 
Amnesic MP & 0.24 & 1.22 & 0.19  \\
Amnesic LEACE & 0.11 & 0.55 & 1.17 \\
\hline
Gold labels INLP & 87.06 & 92.23 & 96.56 \\  
Gold labels MP & 96.75 & 96.81 & 96.8 \\
Gold labels LEACE & 96.58 & 96.83 & 96.79 \\
\hline
Accuracy INLP $\Delta$ & 81.14 & 87.28 & 96.07 \\
Accuracy MP $\Delta$ & 95.51 & 95.59 & \textbf{96.61} \\
Accuracy LEACE $\Delta$ & \textbf{96.47} & \textbf{96.28} & 95.62 \\
\hline
\end{tabular}
\caption{Selectivity Accuracy $\Delta$ show the highest change between embeddings without the information and embeddings with returned gold label information.}
\label{table:selectivity}
\end{table}

\section{Conclusion}

We presented a systematic comparison of three different information removal methods for amnesic probing. We hypothesized that the potential of amnesic probing as a model analysis method increases when using a more precise removal method, specifically MP or LEACE instead of INLP. Our results confirm the hypothesis; MP and LEACE outperform INLP on the information and selectivity control tests. In particular, in two out of three cases, random projections had a bigger impact on the main task than the model modified by INLP, which did not happen when MP or LEACE was used (information control).
In all three cases, MP or LEACE-modified models recover performance better than INLP when explicit gold information about the target property is added (selectivity control). Information and selectivity control provide evidence that the drop in performance on the main task is indeed caused by the target information missing rather than by other artifacts of the modification. As such, our results show that MP and LEACE are more precise methods for removing information, which directly increases the potential of Amnesic Probing. 
MP and LEACE yield highly comparable results and we provide easy-to-use code for both approaches. We recommend to first use MP since this is more efficient and adding experiments with LEACE when checks and balances on the MP results call for it.

% \begin{itemize}
% \item Hypothesized that potential of Amnesic Probing increases when using a more precise method for removing information, specifically MP instead of INLP.
% \item Results confirm our hypothesis: outperform on information and selectivity test. In particular, in two out of three cases, random projections had a bigger impact on the main task than the model where INLP removed results. This indicates... Also recover better: therefore clearer that the impact on main task results 
% \item increases potential of Amnesic Probing
% \end{itemize}

 %\paragraph{Additional Insights on the modified space} The results above confirm that MP indeed removes information in a more targeted way than INLP. We carry out two additional checks to compare the extent to which each approach modifies the original space, namely, (1) the change in matrix rank \cite{Strang2015IntroLinear} and (2) cosine similarity between layer from the original model to the modified model. This comparison confirms that INLP has a much larger impact on the overall model compared to MP. First, the matrix rank change scores are more than 10 times higher across all tasks. Second, MP shows a relatively high overall cosine similarity ranging from 0.80 to 0.91 depending on the linguistic property that is removed. The range for INLP lies between 0.31 and 0.37. A detailed description of these metrics and the results can be found in Appendix~\ref{app:modelrankcosine}.

 \newpage

 \section{Limitations}

\paragraph{Sensitivity to outliers} MP calculates the mean target property vector for each class and uses the difference between these means to define the projection directions. 
As such, it addressed the problem of unbalanced classes (which is a problem for INLP).
MP is, however, sensitive to outliers, as the mean of a class can be affected significantly by extreme values, shifting the projection direction in a way that does not accurately reflect the central tendency of the class distribution. If there are a small number of outliers in a class, they can disproportionately influence the calculated mean, resulting in projection directions that are misaligned with the true data distribution. Because the projection direction is determined by class means, an outlier can cause MP to remove information that is not representative of the class structure. This can result in incomplete removal of the target attribute or unnecessary distortion of the embedding space. Therefore, the control tests of the amnesic probing framework (conducted in this paper) should be standardly included in any Amnesic Probing experiment.

If control tests do point to a problem with outliers, this limitation is manageable and can potentially be addressed by outlier detection methods and does not outweigh the usefulness of MP. For example, we could identify and remove outliers before computing the class means, which could improve the robustness of MP. 

%MP is computationally efficient, causes minimal distortion of the embedding space compared to iterative methods such as INLP, and is easy to interpret. MP remains a practical and computationally efficient alternative to iterative methods, making it a valuable tool for linear concept erasure and interpretability tasks.

% An important limitation of the MP removal method is that it is sensitive to the distribution of the probing data. While it addresses the problem of unbalanced classes (a problem of INLP), it is still sensitive to data with a lot of outliers. Therefore, the control tests of the Amnesic Probing framework (carried out in this paper) should standardly be included in any Amnesic Probing experiment. Outlier analysis methods could be considered to support the method. 

 % MP calculates the mean for each class to determine projection directions, making it inherently sensitive to the distribution of data points within each class. Outliers in a class can affect the mean, resulting in projection directions that do not fully represent the true structure of the class. This can lead to suboptimal removal of target information, which is a recognised limitation of the MP method. However, this limitation is manageable and can potentially be addressed by outlier detection methods, and does not outweigh the usefulness of MP. MP is computationally efficient, causes minimal distortion of the embedding space compared to iterative methods such as INLP, and is easy to interpret.

 \paragraph{Different Results in Reproduction} Our reproduced results differ from those reported in \newcite{Elazar:etal:2021}. The differences in the results for dep and f-pos are largely due to random factors in the INLP method (initialization of the weights, different seeds). These random factors determine the number of iterations INLP needs to remove a property. We observe that INLP needed a higher number of iterations in our experiments than in the original experiments. The results we reproduced for \textit{cpos} and \textit{dep} are relatively close to the original results: the \textit{cpos} results tell the same story. Looking at it from the three dimensions of reproduction introduced by \newcite{cohen2018three}: the numbers differ (dimension 1), but the overall findings (dimension 2) and conclusions (dimension 3) are the same for these two properties. In the case of \textit{dep}, \newcite{Elazar:etal:2021} report around 12\% accuracy with random projections and 7\% for INLP amnesic, whereas we report around 5\% for random and 9\% for INLP amnesic: even though our experiments show a bigger loss for random compared to amnesic, the drop is extremely large leading to very poorly performing models in all cases. We therefore also consider these results reproduced according to the second and third dimension of \newcite{cohen2018three}. 
 
 The reproduction difference is biggest for \textit{fpos}. Here, random projections still perform at 56.47 for \newcite{Elazar:etal:2021} with INLP amnesic dropping to 12.3. In our case, \textit{fpos} drops to 4.8, which is more than the drop caused by INLP amnesic, which drops to 14.6\%. This means that while \newcite{Elazar:etal:2021} show an impact of \textit{fpos} on language modeling, we do not reproduce this finding. We have added the original results from \newcite{Elazar:etal:2021} to the results in Table~\ref{table:lm_non_masked}. Note that MP (and LEACE) also reveal a more precise measurement of the impact of \textit{fpos} compared to the original results in \newcite{Elazar:etal:2021}. We can observe that model performance barely changes when applying as many random projections as projections are required in MP/LEACE. %Applying the same number of random projections as these methods require barely impacts the model performance.

 \paragraph{General Limited Scope and Expanded Motivation for Focusing on the 3 Tasks} It is important to consider that the experiments carried out in this paper included only a limited set of linguistic properties and models. It is possible that the method behaves differently on other properties or models. Note, however, that including the three other control tasks from the original paper by \newcite{Elazar:etal:2021} could not contradict the claims in this paper. INLP does not work for these tasks. Most likely, MP and LEACE do not work (well) either, but they could not do significantly worse. In other words, INLP cannot outperform MP or LEACE here. The chances of MP or LEACE doing better than INLP are very small, therefore we highlight that it is unlikely that adding these experiments would lead to new insights. We still considered adding these three tasks for the sake of completeness, but the data was not available and the authors of the original paper could not share it with us. This made it impossible to reproduce their original results. The combination of the limited insights we could gain from it and the problems around the data led to the decision to not include these experiments in our paper.

 \paragraph{Runs} The results reported in the paper are based on a single run. We did test whether they are stable across multiple runs, notably for INLP and check whether they were influenced by batch size. We got similar results in all runs. Due to computational constraints, we were unable to consistently carry out the same number of runs for each experiment. Since our results do not depend on minor differences (i.e.\ they would not tell a different story based on multiple runs), we decided to report results based on one run.
\section*{Acknowledgements}
This research is supported by the Dutch National Science Organisation (NWO) through the project InDeep: Interpreting Deep Learning Models for Text and Sound (NWA.1292.19.399). We would like to thank the anonymous reviewers whose feedback helped improve this paper. All remaining errors are our own.

\bibliography{custom}

\newpage
\appendix

% 1 batch 3 epochs
% normal:
% Micro precision: 77.90
% Micro recall:    65.37
% Micro F1:        71.09

% mp dep removed: (non-masked projection)
% Micro precision: 76.56
% Micro recall:    65.46
% Macro F1:        70.58

% random:
% Micro precision: 66.83
% Micro recall:    53.36
% Macro F1:        59.34

% Discussion
% random results are still high, it is not clear what it happening
% in the probing
% for amnesic probing relatively straight setup

\section{INLP and MP Differences}
\label{appendix:inlp_mp_differences}

Both MP and INLP reply on the concept of projection from linear algebra. Projection involves finding the closest vector for a given subspace. Geometrically, projection can be thought of as the shadow of the target vector on a subspace. The steps behind removing the target properties from the embedding using projection methods are as follows:

\begin{enumerate}
    \item Find the direction for the projection onto the subspace\label{list:direction}
    \begin{enumerate}
        \item use weights from a linear classifier (INLP)
        \item use the difference between mean class vector and the mean of the remaining vectors for all classes (MP)
    \end{enumerate}
    \item Get the orthogonal projection matrix over the rowspace, which is equivalent to obtaining a nullspace \cite{ravfogel-etal-2020-null}\label{list:orthogonal_projection}
    \begin{enumerate}
        \item Use the formula from Adi Ben–Israel \cite{BenIsrael2013Projectors} to get orthogonal projection to the intersection of nullspaces while running multiple iterations (INLP)
        \item Use the orthogonal projection over the rowspace once (MP)
    \end{enumerate}
    \item Get a projection onto the orthogonal complement\label{list:orthogonal_complement}
\end{enumerate}

Point~\ref{list:direction} refers to the fact that a direction or set of directions is needed to perform a projection. The directions define the subspace on which the projection is performed. The goal of both methods is to find the directions that most accurately capture the information about the target property classes. Point~\ref{list:orthogonal_projection} talks about getting an orthogonal projection using the directions defined in the previous step. Such a projection \textit{P}, when applied to the embedding, should ensure that the classes of the target property are no longer linearly separable. Point~\ref{list:orthogonal_complement} refers to the last step, where subtracting the intersection of the nullspaces from the identity matrix gives the projection to the union of the rowspaces of each direction \cite{ravfogel-etal-2020-null}.

\begin{figure}[ht]
\includegraphics[width=8cm]{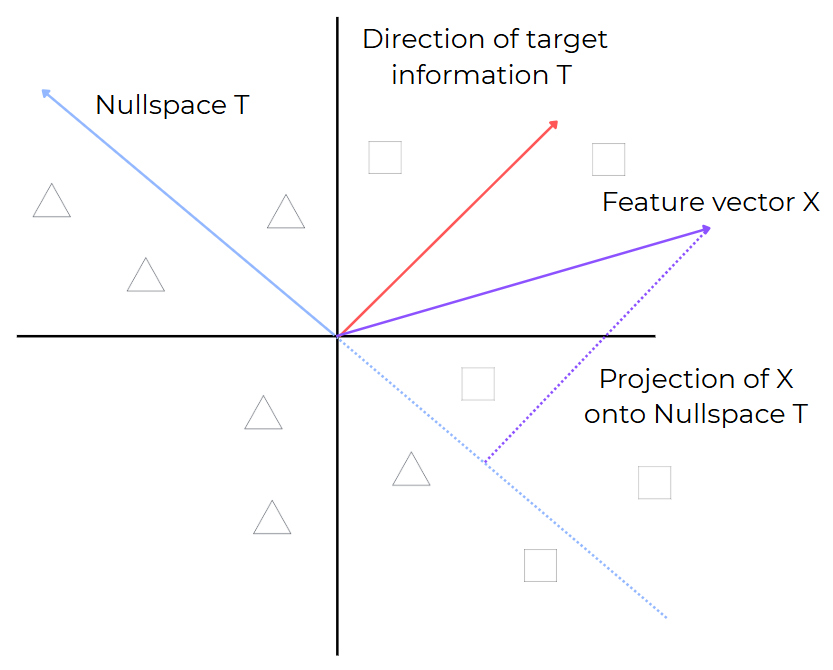}
\caption{A simplified visualisation of the removal of target information using linear projection.}
\label{fig:mp_visualized}
\end{figure}

Figure~\ref{fig:mp_visualized} provides a visualisation of the step of removing target information.

An overview of the comparison of the two methods is shown in Table~\ref{table:mp_inlp_differences}.

\begin{table}[htbp]
\centering
\setlength{\tabcolsep}{3pt} 
\renewcommand{\arraystretch}{1.3}
% \begin{tabular}{ c | c | c }
\begin{tabular}{p{0.5\linewidth} p{0.5\linewidth}} 
\hline

\multirow{2}{*}{goal} & similar class distributions (MP)\\
 & linear guarding (INLP)\\

\hline
\multirow{2}{*}{multiple iterations}& no (MP) \\
 & yes (INLP) \\
\hline
\multirow{2}{*}{directions source}  & means of the classes (MP) \\
 & classifier's weights (INLP)\\
\hline
\multirow{2}{*}{no. directions removed}  & equal to the no. classes (MP) \\
 & minimum the no. classes (INLP) \\
\hline
\multirow{2}{*}{orthogonal projection} & yes (MP)\\
 & yes (INLP)\\
\hline
\end{tabular}
\caption{Similarities and differences between MP and INLP.}
\label{table:mp_inlp_differences}
\end{table}

\paragraph{Why MP removes information in a more targeted way compared to INLP} Amnesic probing aims to determine whether a model actually uses a particular property by removing it and observing the change in performance. If the embedding space is significantly altered beyond simply removing the target property, the results can be misleading - suggesting that the model relied on the property, when in fact the performance drop could be due to broader distortions introduced by the projection method. While both MP and INLP aim to remove certain properties from embedding space, MP is a more targeted and precise approach. MP applies a single projection, whereas INLP requires multiple iterations. This fundamental difference results in several advantages for MP in terms of stability, efficiency and reduced damage to the general embedding space.

INLP iteratively identifies and removes directions in the embedding space that encode the target property. However, after several iterations, the projections become increasingly random and less targeted, leading to unintended alterations in the embedding space \cite{Haghighatkhah:etal:2022}. This phenomenon results in greater distortion of non-target information, making it difficult to isolate the effect of the target property removal. \newcite{belrose2024leace} also highlight this risk of ``collateral damage" in INLP, noting that as the number of projections increases, the embedding space undergoes cumulative distortions that are not necessarily related to the removal of target information.

Studies comparing INLP and MP show that MP alters the embedding space less than INLP while achieving comparable target property removal. For example, KL divergence measurements indicate that MP produces a more stable token distribution than INLP. In addition, probing accuracy tests confirm that MP successfully removes the target property while better preserving the original space. INLP also typically requires 10-15 iterations to make the target information linearly inseparable. MP achieves the same level of separability with a single projection, which means less computation, less distortion, and a more precise modification of space. Because of these advantages, MP is recommended over INLP for applications that require precise information removal.

\begin{figure}[ht]
\includegraphics[width=8cm]{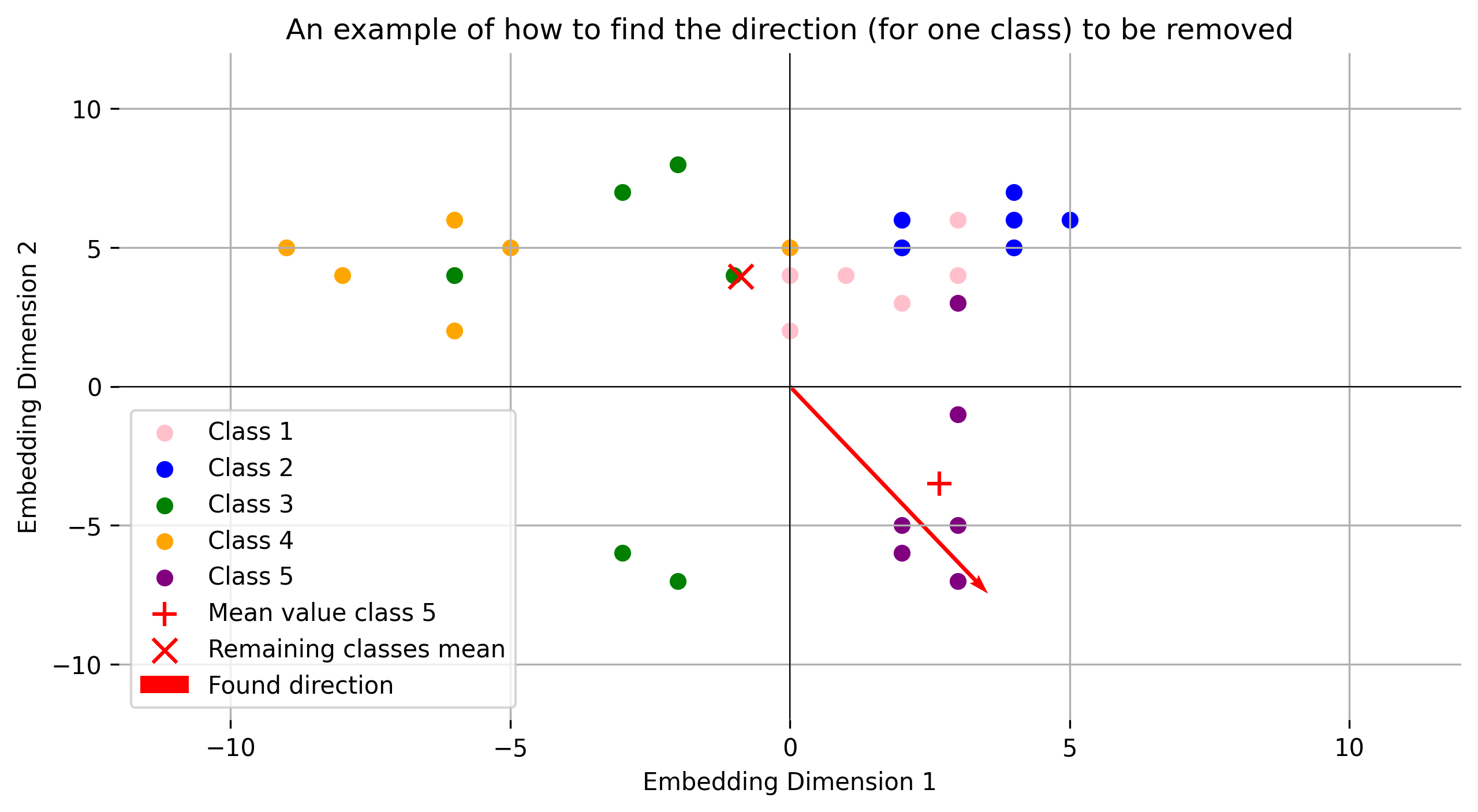}
\caption{Intuitive example of how to identify a target direction for removal in two-dimensional space.}
\label{fig:direction}
\end{figure}

\begin{figure}[ht]
\includegraphics[width=8cm]{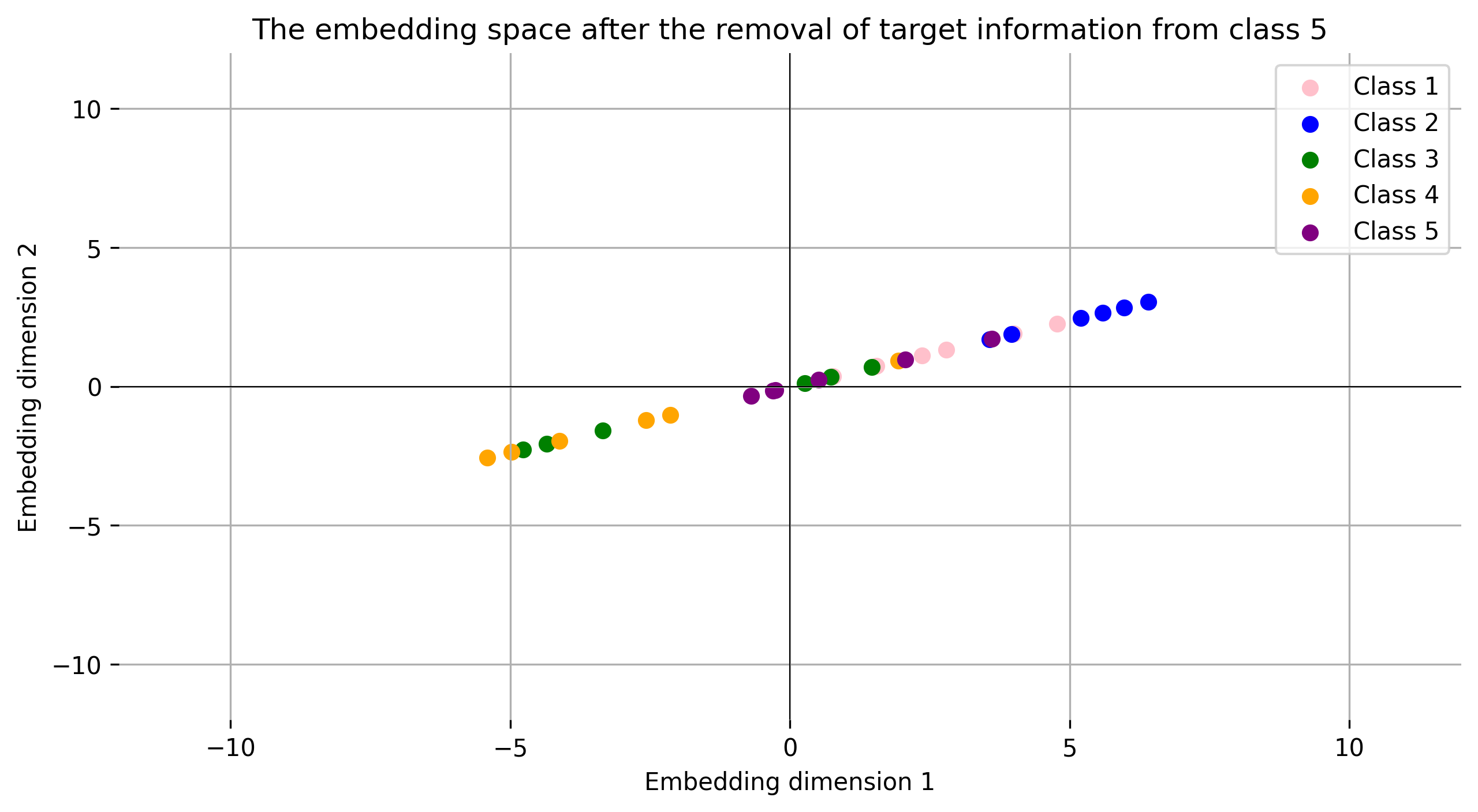}
\caption{Intuitive example of how the embedding space changes after the removal of one target direction in two-dimensional space.}
\label{fig:after_removal}
\end{figure}

Figures~\ref{fig:direction} (before) and~\ref{fig:after_removal} (after) provide an illustration of how target information is identified and removed through projection.

% MP example for one direction removal

\section{Visualisation of the Amnesic Probing Experimental Set-up}

\begin{figure*}[ht]
\begin{center}
\includegraphics[width=16cm]{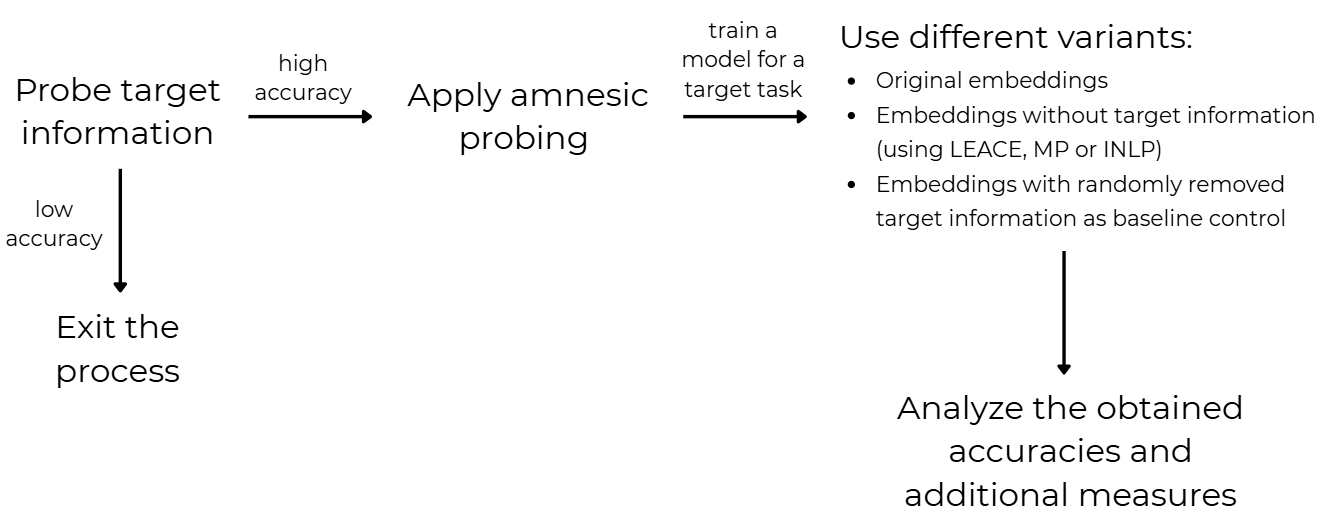}
\caption{The conceptual visualisation of the amnesic probing framework.}
\label{fig:framework}
\end{center}
\end{figure*}

A schematic representation of the steps involved in our experimental setup can be found in Figure~\ref{fig:framework}.

% MP visualization

\section{Results for Masked Setup}
\label{appendix:masked_setup}

\subsection{Probing Evaluation for Masked BERT}

The results of the probing evaluation for masked BERT can be found in Table \ref{table:probing_masked}. 
\begin{table}[htbp]
\centering
% \scalebox{1.3}{
\setlength{\tabcolsep}{6pt} 
\renewcommand{\arraystretch}{1.3}
\begin{tabular}{ c c c c c}
\hline
&  & \textit{dep} & \textit{f-pos} & \textit{c-pos} \\
\hline
& Vanilla & 72.87 & 77.81 & 84.44 \\ 
& Vanilla LEACE & 83.88 & 77.9 &72.39 \\
& Amnesic INLP & \textbf{11.07} & \textbf{11.51} & 27.5 \\ 
& Amnesic MP & 15.44 & 15.94 & \textbf{25.7} \\ 
& Amnesic LEACE & 25.69 & 16.13 & 16.14 \\
\hline
\end{tabular}
% }
\caption{Results for probing for masked BERT encoding. In bold the lowest values achieved per property.}
\label{table:probing_masked}
\end{table}

\subsection{Evaluation Controls for Masked BERT}

The results of the evaluation controls for masked BERT can be found in Table \ref{table:embedding_masked}. 

\begin{table}[htbp]
\centering
\setlength{\tabcolsep}{6pt} 
\renewcommand{\arraystretch}{1.6}
\begin{tabular}{ c c c c c}
\hline

& \textit{dep} & \textit{f-pos} & \textit{c-pos} \\
\hline
Dir. removed INLP & \textbf{738} & \textbf{900} & \textbf{240} \\
Dir. removed MP & 41 & 45 & 12 \\
Dir. removed LEACE & 41 & 45 & 12 \\
\hline
Rank INLP $\Delta$ & \textbf{425} & \textbf{440} & \textbf{210} \\ 
Rank MP $\Delta$  & 33 & 37 & 4 \\ 
Rank LEACE $\Delta$  & 32 & 34 & 10 \\ 
\hline
Cosine random INLP & 0.19 & -0.03 & 0.83  \\
Cosine random MP & 0.97 & 0.97 & 0.99 \\
Cosine random LEACE & 0.97 & 0.97 & 0.99 \\
\hline
Cosine similarity INLP & 0.50 & 0.47 & 0.75 \\
Cosine similarity MP  & 0.81 & 0.78 & 0.88 \\
Cosine similarity LEACE  & \textbf{0.89} & \textbf{0.86} & \textbf{0.93} \\

\end{tabular}
\caption{Embedding evaluation for masked BERT. The original number of columns in the embedding is $768$. The rank of the matrix for the test set before any modification is $760$. Values in bold for Properties indicate the largest change amount of Directions removed and the largest change in the rank between the same properties. Values in bold for Cosine similarity indicate which information removal resulted in the highest cosine similarity between the original embedding and the modified one between the same properties.}
\label{table:embedding_masked}
\end{table}

\subsection{LM Task for Masked BERT}

The results of the LM task for masked BERT can be found in Table \ref{table:lm_masked}. 

\begin{table}[htbp]
\centering
% \scalebox{1.3}{
\setlength{\tabcolsep}{6pt} 
\renewcommand{\arraystretch}{1.6}
\begin{tabular}{ c c c c c }

\hline
 &  & \textit{dep} & \textit{f-pos} & \textit{c-pos} \\
\hline
\multirow{4}{*}{Acc} & Vanilla & 56.98 & 56.98 & 56.98 \\
& Random INLP & 9.24 & 4.67 & 54.29 \\
& Random MP & 56.70 & 56.69 & 56.87 \\ 
& Random LEACE & 56.54 & 56.58 & 56.74 \\ 
\hline
& Amnesic INLP & 16.96 & 13.94 & \textbf{33.29}  \\ 
& Amnesic MP & \textbf{44.61} & 
\textbf{34.20} & \textbf{48.56} \\

& Amnesic LEACE & \textbf{48.20} & 
\textbf{37.60} & \textbf{50.44} \\
\hline
\multirow{2}{*}{$D_{kl}$} & Random INLP & 7.06 & 7.77 & 0.49 \\
& Random MP & 0.03 & 0.04 & 0.01\\ 
& Random LEACE & 0.03 & 0.04 & 0.01\\ 
\hline
& Amnesic INLP & 5.75 & 6.08 & 3.1  \\ 
& Amnesic MP &  \textbf{1.45} & \textbf{2.09} & \textbf{0.86} \\
& Amnesic LEACE &  \textbf{0.74} & \textbf{1.37} & \textbf{0.50} \\

\hline
\end{tabular}
% }
\caption{LM task for masked BERT encoding. Values in bold for LM-Acc Amnesic show which values obtained the accuracy lower than their Random and Dropout baselines. Values in bold for the $LM-D_{kl}$ indicate which properties with which method resulted in the lowest change of the general distribution of the tokens.}
\label{table:lm_masked}
\end{table}

\subsection{Selectivity Results for Masked BERT}

The results of the selectivity control for masked BERT can be found in Table \ref{table:lm_masked}. 

\begin{table}[htbp]
\centering
% \scalebox{1.3}{
\setlength{\tabcolsep}{6pt} 
\renewcommand{\arraystretch}{1.6}
\begin{tabular}{ c c c c c }
\hline
& \textit{dep} & \textit{f-pos} & \textit{c-pos} \\
\hline
Amnesic INLP & 4.1 & 5.43 & 2.96 \\  
Amnesic MP & 1.63 & 1.7 &  1.54 \\
Amnesic LEACE & 0.70 & 1.59 & 0.22 \\
\hline
Gold labels INLP & 61.95 & 63.97 & 64.63 \\ 
Gold labels MP & 64.72 & 66.95 & 65.09 \\
Gold labels LEACE & 64.41 & 66.80 & 64.78 \\
\hline
Accuracy INLP $\Delta$ & 57.85 & 58.54 & 61.67 \\
Accuracy MP $\Delta$ & 63.09 & \textbf{65.25} & 63.55 \\
Accuracy LEACE $\Delta$ & \textbf{63.71} & 65.21 & \textbf{64.56} \\
\hline
\end{tabular}
% }
\caption{LM task for masked BERT encoding. Values in bold for the difference in accuracy for Selectivity show the highest change between embeddings without the information and embeddings with returned gold label information.}
\label{table:lm_masked_selectivity}
\end{table}

\section{Experimental Details}
\label{appendix:experiments}

\subsection{Data details}

The details of our dataset can be found in Table \ref{table:sent_and_token_counts}. 
\begin{table}[htbp]
\centering
% \scalebox{1.3}{
\setlength{\tabcolsep}{6pt} 
\renewcommand{\arraystretch}{1.6}
\begin{tabular}{c c | c}
\hline
& \textit{no. sentences} & \textit{no. tokens} \\
\hline
train set & 39832 & 1113133 \\ 
test set & 1700 & 47095 \\ 
\hline
% \multirow{2}{*}{SRL task} & train set & 30000 & 813725 \\ 
%  & test set & 1500 & 32998  \\ 

\end{tabular} 
% }
\caption{Number of sentences and tokens used per Language Modeling (LM) task.}
\label{table:sent_and_token_counts}
\end{table}

\subsection{Model details}

The details of our models can be found in Table \ref{table:model_details}. 

\begin{table}[ht]
\centering
% \scalebox{1.3}{
\setlength{\tabcolsep}{6pt} 
\renewcommand{\arraystretch}{1.6}
\begin{tabular}{p{0.2\linewidth} p{0.2\linewidth} p{0.2\linewidth} p{0.2\linewidth}} %{ c c c c}
\hline
 & \textit{model} & \textit{Library} \\
\hline
Probing & Linear classifier with SGD training & Sklearn \\
\hline
Task (LM) & Linear transformation $y=xA^{T}+b$ and Sequential container & Pytorch  \\ 
\hline
% Task (SRL) & Linear classifier with SGD trainining & loss = log, \newline warm\_start = True, \newline partial fit with batch size= 100000 & Sklearn \\
% \hline
Selectivity (LM) & Linear transformation $y=xA^{T}+b$ &  Pytorch \\ 
\hline
Rank of matrix & Matrix rank using SVD method & Numpy  \\
\hline
Cosine Similarity &  & Sklearn \\

\end{tabular}
% }
\caption{Implementation details}
\label{table:model_details}
\end{table}

\section{The Overall Impact on the Model}\label{app:modelrankcosine}

In this section, we describe the two additional metrics we used to study the impact of INLP, MP and LEACE on the model. The dimensions of a matrix do not necessarily indicate the true size of a linear system. If there are duplicate rows, or rows that are combinations of other rows, then, in the reduced echelon form, they are all zeros. To better understand the peculiarities of a given matrix, instead of referring to its dimensions, we include in the experiments the measure of matrix rank \cite{Strang2015IntroLinear}. Matrix rank indicates a minimum number of linearly independent rows and columns in a given matrix. It does not give a direct measure of the amount of information present in the data, but it can give an indication of the decrease in linearly independent columns. This measure can provide additional information on how much the rank decreases when LEACE, MP or INLP is used. A larger change can potentially indicate a larger modification of the original data.

To further account for the difference in the embedding that LEACE, MP and INLP perform, we also include a cosine similarity measure. Cosine similarity is used to determine the degree of similarity between two vectors \cite{Lahitani2016CosineMeausre}.\footnote{It is calculated by dividing the vectors' dot product by the product of their lengths. The resulting value ranges from -1 to 1, where a value of 1 indicates the maximum similarity, 0 indicates no similarity, and -1 indicates the maximum dissimilarity.} This measure can be a good additional check of how much the LEACE, MP and INLP methods modify the original data while removing target information. If one method correctly identifies and eliminates the target property, but modifies the original space less than the other method, then that method should be preferred for the Amnesic Probing analysis. The cosine similarity between the test data before and after information removal is calculated by first taking the cosine similarities between all corresponding rows from the original and modified data and then averaging the result. 

In Table \ref{table:embedding_non_masked} the number of removed directions, the change in the rank and the cosine similarity between the original and the modified data are given for each method and linguistic property. The Rank $\Delta$ gives an estimate of how much the rank of the matrix has changed after the information removal. 

It can be observed that the number of directions removed and the change in the rank is the largest for INLP. It is also worth noting that as the number of directions removed increases, so does the change in the rank of the matrix, meaning that less linearly independent information is encoded. One outstanding value for the change of the rank for MP is $-17$ for the \textit{c-pos} property.\footnote{This value indicates that after the projection the rank of matrix increased by $17$. There is no clear answer as to why this opposite trend occurred. One of the possible explanations is that because MP and LEACE only removed $12$ directions for the \textit{c-pos} property, instead of losing linearly independent rows and columns, we actually gained some because the projection removed some noise.}

Looking at the cosine similarity results, we can see that MP and LEACE modify the embedding space in such a way that the similarity between original and modified data is still quite high, with values between $0.80$-$0.91$ for MP and $0.89$-$0.95$ for LEACE depending on the linguistic property. In the case of INLP, the similarity with the original embedding after information removal is much lower for two out of three properties, for the linguistic properties \textit{dep} and \textit{f-pos} it is $0.31$ and $0.37$ respectively. The \textit{c-pos} property is an exception for INLP, as it achieves a cosine similarity of $0.80$, probably due to the lower amount of direction removed compared to the other properties. In Table \ref{table:embedding_non_masked} there is additional indication of Cosine similarity when random directions are removed. In row Rand we can see the relation between the amount of directions removed and the cosine similarity value. The fewer directions are removed, the more the modified embedding resembles the original one. 

\begin{table}[htbp]
\centering
\setlength{\tabcolsep}{6pt} 
\renewcommand{\arraystretch}{1.3}
\begin{tabular}{ c c c c c }
\hline
& \textit{dep} & \textit{f-pos} & \textit{c-pos} \\ 
\hline
Dir. removed INLP & \textbf{779} & \textbf{765} & \textbf{240} \\
Dir. removed MP  & 41 & 45 & 12 \\
Dir. removed LEACE & 41 & 45 & 12 \\
\hline
Rank INLP $\Delta$  & \textbf{629} & \textbf{617} & \textbf{198} \\
Rank MP $\Delta$ & 12 & 16 & -17 \\ 
Rank LEACE $\Delta$ & 31 & 34 & 9 \\
\hline
Cosine random INLP & -0.04 & 0.05 & 0.83 \\
Cosine random MP & 0.97 & 0.96 & 0.99 \\
Cosine random LEACE & 0.97 & 0.99 & 0.97 \\
\hline
Cosine similarity INLP & 0.31 & 0.37 & 0.80 \\
Cosine similarity MP & 0.83 & 0.80 & \textbf{0.91} \\ 
Cosine similarity LEACE & \textbf{0.92} & \textbf{0.95} & 0.89 \\
\hline
\end{tabular}
\caption{Embedding evaluation for non-masked BERT. The original number of columns in the embedding is $768$. The rank of the matrix for the test set before any modification is $739$. Values in bold for Properties indicate the largest amount of directions removed and the largest change in the rank between the same properties. Values in bold for Cosine similarity indicate which information removal resulted in the highest cosine similarity between the original embedding and the modified one between the same properties.}
\label{table:embedding_non_masked}
\end{table}

In summary, LEACE, MP and INLP differ in how they modify the general embedding space. INLP removes many more directions and also reduces the rank of the matrix to a greater extent. The cosine similarity between the original data and the data with the target information removed is higher for the MP and LEACE methods. This suggests that MP and LEACE can potentially preserve the general embedding space better while identifying and removing the target linguistic property as well as INLP.

\section{Computing Resources}

The experiments were run on a local laptop with the following specifications: 

\begin{itemize}
    \item Processor: AMD Ryzen™ 7 4800H (8 cores, 16 threads, 2.90-4.20 GHz, 12 MB cache)
    \item Graphics card: NVIDIA GeForce GTX 1660 Ti AMD Radeon™ Graphics
\end{itemize}

It took around 8-12 hours to run selectivity for one case (e.g.\ MP and dependency as target information) and around 4-6 hours to run the rest of the evaluation.

\end{document}